\documentclass[11pt,a4paper]{article}
\usepackage[hyperref]{emnlp2018}

\usepackage{array}
\usepackage{amsfonts}
\usepackage{amsmath}

\usepackage{bbm}
\usepackage{boldline}
\usepackage{bigstrut}
\usepackage{blindtext}
\usepackage{booktabs, siunitx}
\usepackage[labelfont=bf, format=plain, justification=justified, singlelinecheck=false]{caption}
\usepackage{color}
\usepackage{cprotect}
\usepackage{ctable}
\usepackage{dirtytalk}
\usepackage{enumitem}
\usepackage[export]{adjustbox}
\usepackage{float}
\usepackage{graphicx}
\usepackage{latexsym}
\usepackage{mathrsfs}
\usepackage{microtype}
\usepackage{moresize}
\usepackage{multirow}
\usepackage{nccmath}
\usepackage{nicefrac}
\usepackage{pifont}
\usepackage{placeins}
\usepackage{soul}
\usepackage{subcaption}
\usepackage{tabularx}
\usepackage{times}
\usepackage[utf8]{inputenc}
\usepackage{url}
\usepackage{verbatim}
\usepackage{wrapfig, lipsum}

\usepackage[T1]{fontenc}    
\newcolumntype{M}[1]{>{\centering\arraybackslash}m{#1}}

\aclfinalcopy 

\title{Parameter Sharing Methods for \\ Multilingual Self-Attentional Translation Models}

\author{Devendra Singh Sachan \\
  Data Solutions Team \\
  Petuum Inc. \\
  Pittsburgh, USA \\
  {\tt devendra.singh@petuum.com} \\\And
  Graham Neubig \\
  Language Technologies Institute \\
  Carnegie Mellon University \\
  Pittsburgh, USA \\
  {\tt gneubig@cs.cmu.edu} \\}

\date{}
\begin{document}
\maketitle
\begin{abstract}
  In multilingual neural machine translation, it has been shown that sharing a single translation model between multiple languages can achieve competitive performance, sometimes even leading to performance gains over bilingually trained models. However, these improvements are not uniform; often multilingual parameter sharing results in a decrease in accuracy due to translation models not being able to accommodate different languages in their limited parameter space. In this work, we examine parameter sharing techniques that strike a happy medium between full sharing and individual training, specifically focusing on the self-attentional \emph{Transformer} model. We find that the full parameter sharing approach leads to increases in BLEU scores mainly when the target languages are from a similar language family. However, even in the case where target languages are from different families where full parameter sharing leads to a noticeable drop in BLEU scores, our proposed methods for partial sharing of parameters can lead to substantial improvements in translation accuracy.%
\footnote{Data and code of this paper is available at: \href{https://github.com/DevSinghSachan/multilingual\_nmt}{https://github.com/DevSinghSachan/multilingual\_nmt}}
\end{abstract}

\section{Introduction}
\begin{figure}[!ht]
\centering
\begin{minipage}{\linewidth}
  \centering
  \includegraphics[scale=0.9]{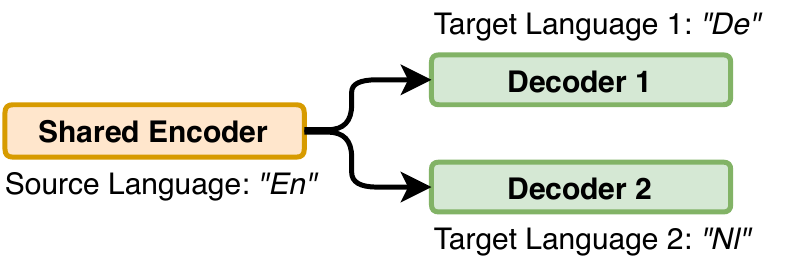}
  \subcaption{Shared encoder, separate decoder~\citep{dong2015multi}.}
  \label{fig:o2m_decoder}
  \medskip
\end{minipage}
\begin{minipage}{\linewidth}
  \centering
  \includegraphics[scale=0.9]{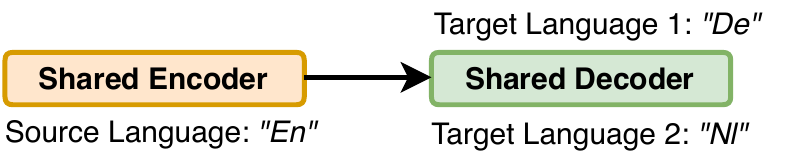}
  \subcaption{Shared encoder and decoder~\citep{johnson2017google}.}
  \label{fig:google_mt}
  \medskip
\end{minipage}
\begin{minipage}{\linewidth}
  \centering
  \includegraphics[scale=0.9]{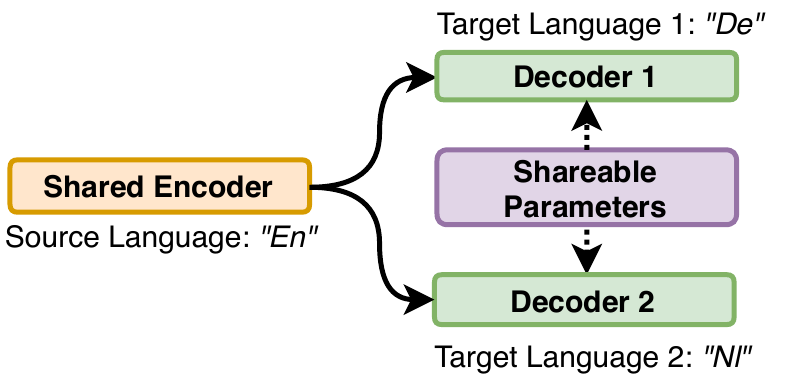}
  \subcaption{Proposed shared decoder with partial parameter sharing.}
  \label{fig:o2m_deocder_partial}
\end{minipage}
\caption{Examples of MTL frameworks for the translation of one source language (for example \emph{\say{En}}) to two target languages (for example \emph{\say{De}}, \emph{\say{Nl}}). The principle remains the same with more than two target languages. Best viewed in color.} 
\label{fig:multi_task}
\end{figure}

Neural machine translation (NMT; \citet{sutskever2014nips,kcho2014emnlp}) is now the de-facto standard in MT research due to its relative simplicity of implementation, ability to perform end-to-end training, and high translation accuracy.
Early approaches to NMT used recurrent neural networks (RNNs), usually LSTMs~\cite{hochreiter1997lstm}, in their encoder and decoder layers, with the addition of an attention mechanism~\cite{bahdanau2014neural,luong2015effective} to focus more on specific encoded source words when deciding the next translation target output. Recently, the NMT research community has been transitioning from RNNs to an alternative method for encoding sentences using self-attention~\cite{vaswani2017attn}, represented by the so-called ``\emph{Transformer}'' model, which both improves the speed of processing sentences on computational hardware such as GPUs due to its lack of recurrence, and achieves impressive results.

In parallel to this transition to self-attentional models, there has also been an active interest in the multilingual training of NMT systems~\cite{firat2016multi,johnson2017google,ha2016toward}.
In contrast to the standard bilingual models, multilingual models follow the multi-task training paradigm~\cite{Caruana1997mtl} where models are \emph{jointly trained} on training data from several language pairs, with some degree of \emph{parameter sharing}.
The objective of this is two-fold:
First, compared to individually training separate models for each language pair of interest, this maintains competitive translation accuracy while reducing the total number of models that need to be stored, a considerable advantage when deploying practical systems.
Second, by utilizing data from multiple language pairs simultaneously, it becomes possible to improve the translation accuracy for each language pair.

In multilingual translation, \emph{one-to-many} translation ---translation from a common source language (for example English) to multiple target languages (for example German and Dutch) --- is considered particularly difficult. Previous multi-task learning (MTL) models for this task broadly consist of two approaches as shown in Figure~\ref{fig:multi_task}: (a) a model with a shared encoder and one decoder per target language (\citet{dong2015multi}, shown in Figure~\ref{fig:o2m_decoder}). This approach has the advantage of being able to model each target separately but comes with the cost of slower training and increased memory requirements.\ (b) a single \emph{unified} model consisting of a shared encoder and a shared decoder for all the language pairs (\citet{johnson2017google}, shown in Figure~\ref{fig:google_mt}). This simple approach is trivially implementable using a standard bilingual translation model and has the advantage of having a constant number of trainable parameters regardless of the number of languages, but has the caveat that the decoder's ability to model multiple languages can be significantly reduced. 

In this paper, we propose a third alternative:
(c) a model with a shared encoder and multiple decoders such that some decoder parameters are shared (shown in Figure~\ref{fig:o2m_deocder_partial}). This hybrid approach combines the advantages from both the approaches mentioned above. It carefully moderates the types of parameters that are shared between the multiple languages to provide the flexibility necessary to decode two different languages, but still shares as many parameters as possible to take advantage of information sharing across multiple languages.
Specifically, we focus on the aforementioned self-attentional Transformer models, with the set of shareable parameters consisting of the various attention weights, linear layer weights, or embedding weights contained therein.
The \emph{full sharing} and \emph{no sharing} of decoder parameters used in previous work are special cases (refer to Section~\ref{subsec:proposed_method} for a detailed description).

To empirically examine the utility of this approach, we examine the case of translation from a common source language to multiple target languages, where the target languages can be either related or unrelated.
Our work reveals that while full parameter sharing works reasonably well when using target languages from the same family, partial parameter sharing is essential to achieve the best accuracy when translating into multiple distant languages.

\begin{figure}[!ht]
\centering
\includegraphics[max width=\linewidth, scale=0.87]{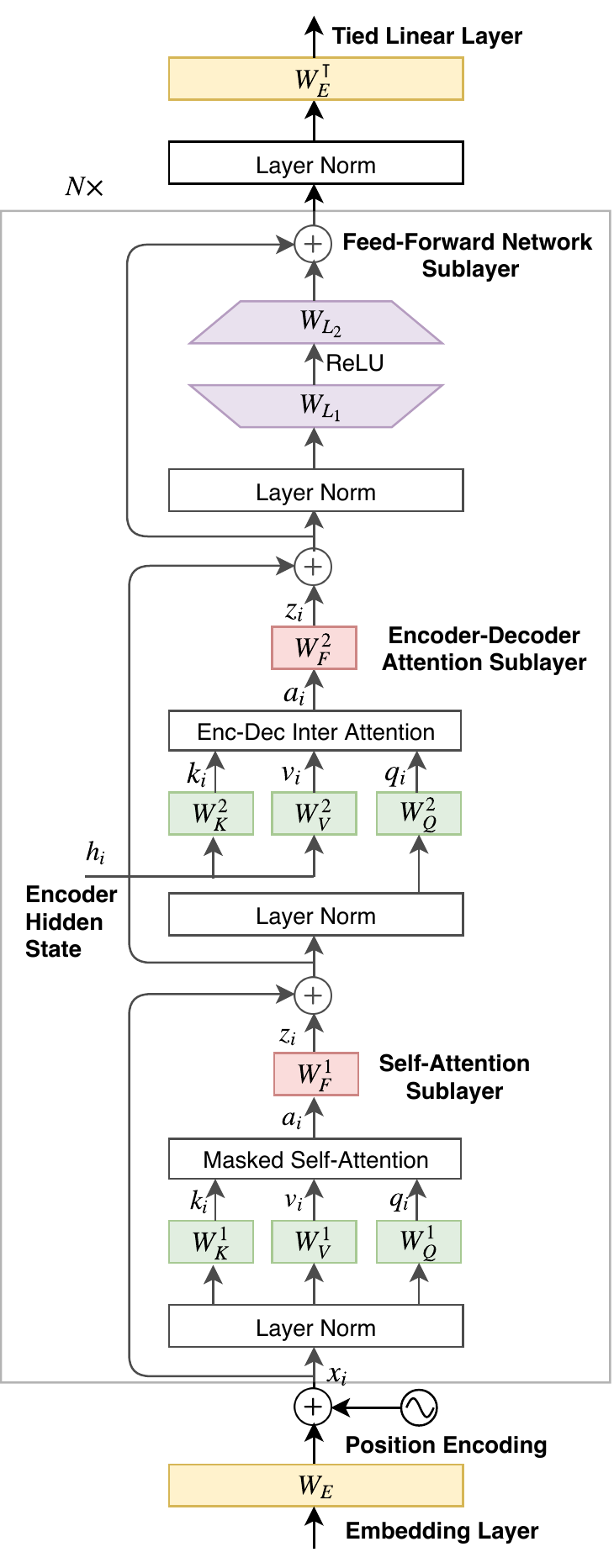}
\caption{Block diagram illustrating the Transformer decoder's shareable parameters (in color) that includes embedding layer weights ($\boldsymbol{W_E}$), tied linear layer weights ($\boldsymbol{W_E^\mathsf{T}}$), transformation weights as a part of self-attention ($\boldsymbol{W_{\textit{K}}^1}, \boldsymbol{W_{\textit{V}}^1}, \boldsymbol{W_{\textit{Q}}^1}, \boldsymbol{W_{\textit{F}}^1}$), encoder-decoder attention ($\boldsymbol{W_{\textit{K}}^2}, \boldsymbol{W_{\textit{V}}^2}, \boldsymbol{W_{\textit{Q}}^2}, \boldsymbol{W_{\textit{F}}^2}$), and feed-forward network ($\boldsymbol{W_{\textit{L}_1}}$, $\boldsymbol{W_{\textit{L}_2}}$) sublayers. Best viewed in color.}
\label{fig:tf_decoder}
\end{figure}

\section{Method}	\label{sec:methods}
In this section, we will first briefly describe the key elements of the Transformer model followed by our proposed approach of parameter sharing. 

\subsection{Transformer Architecture}  \label{subsec:Model}
 
As is common in sequence-to-sequence (\emph{seq2seq}) models for NMT, the self-attentional Transformer model~(Figure~\ref{fig:tf_decoder}; \citet{vaswani2017attn}) consists of an embedding layer, multiple encoder-decoder layers, and an output generation layer. Each encoder layer consists of two sublayers in sequence: self-attentional and feed-forward networks. Each decoder layer consists of three sublayers: masked self-attention, encoder-decoder attention, and feed-forward networks. The core building blocks in all these layers consist of different sets of weight matrices that compute affine transforms.

First, an embedding layer obtains the source and target word vectors from the input words: $\boldsymbol{W_{\textit{E}}}\in\mathbb{R}^{d_{\textit{m}}\times\textit{V}}$, where $d_{\textit{m}}$ is model size, and $\textit{V}$ is vocabulary size.
After the embedding lookup step, word vectors are multiplied by a scaling factor of $\sqrt{d_{\textit{m}}}$.
To capture the relative position of a word in the input sequence, \emph{position encodings} defined in terms of sinusoids of different frequencies are added to the scaled word vectors of the source and target. 

\begin{figure*}[!ht]
\centering
\includegraphics[max width=\linewidth, scale=0.8]{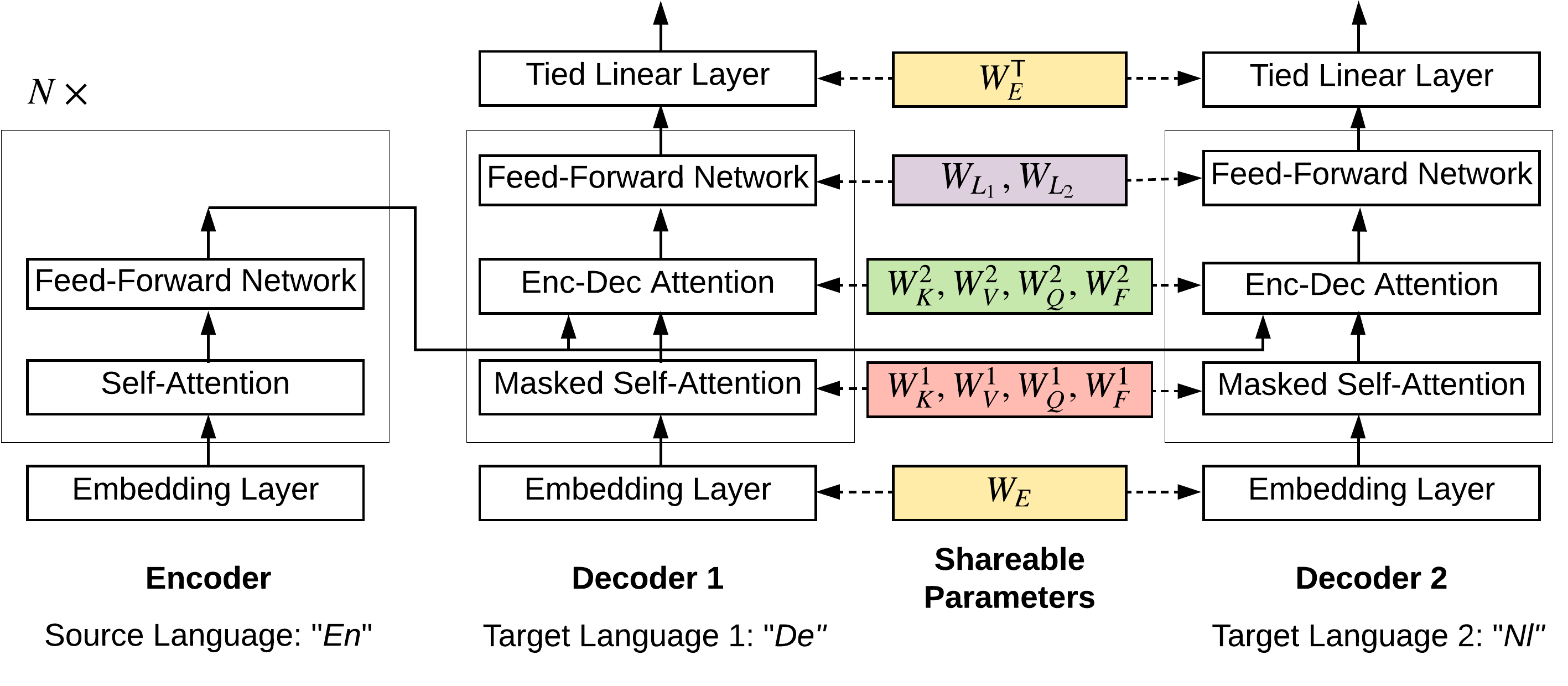}
\caption{Block diagram illustrating our MTL approach for \emph{one-to-many} multilingual translation task that is based on the partial sharing of parameters between the multiple decoders. Best viewed in color.}
\label{fig:pshare}
\end{figure*}

The encoder layer maps the input word vectors to continuous hidden state representations. As mentioned earlier, it consists of two sublayers. The first sublayer performs \emph{multi-head dot-product self-attention}.
In the single-head case, defining the input to the sublayer as $x = (x_1,\ldots,x_{\textit{T}})$ and the output as $z = (z_1,\ldots,z_{\textit{T}})$, where $x_i, z_i \in \mathbb{R}^{d_{\textit{m}}}$, the input is linearly transformed to obtain key ($k_i$), value ($v_i$), and query ($q_i$) vectors
\begin{align*}
k_i=x_i \boldsymbol{W_{\textit{K}}}, 
v_i=x_i \boldsymbol{W_{\textit{V}}}, 
q_i=x_i \boldsymbol{W_{\textit{Q}}}.
\end{align*}
Next, similarity scores ($e_{ij}$) between query and key vectors are computed by performing a scaled dot-product
\begin{align*}
e_{ij} = \frac{1}{\sqrt{d_{\textit{m}}}}q_i k_j^{\textit{T}}.
\end{align*}
Next, attention coefficients ($\alpha_{ij}$) are computed by applying softmax function over these similarity values.
\begin{align*}
\alpha_{ij} = \frac{\exp{e_{ij}}}{\sum_{l=1}^{\textit{T}}\exp{e_{il}}}
\end{align*}
Self-attention output ($z_i$) is computed by the convex combination of attention weights with value vectors followed by a linear transformation
\begin{align*}
z_i = (\sum_{j=1}^T\alpha_{ij}v_j) \boldsymbol{W_{\textit{F}}}.
\end{align*}
In the above equations, $\boldsymbol{W_{\textit{K}}}, \boldsymbol{W_{\textit{V}}}, 
\boldsymbol{W_{\textit{Q}}}, 
\boldsymbol{W_{\textit{F}}}$ are learnable transformation matrices of shape $\mathbb{R}^{d_{\textit{m}}\times{d_\textit{m}}}$. 
To extend to multi-head attention ($\ell$), one can split the key, value, and query vectors into $\ell$ vectors, perform the attention computation in parallel for each of the $\ell$ vectors followed by concatenating before the final linear transformation by $\boldsymbol{W_{\textit{F}}}$.
The second sublayer consists of a two-layer deep \emph{position-wise feed-forward network} ($\mathrm{FFN}$) with ReLU activation~\cite{glorot2011deep}.
\begin{align*}
\mathrm{FFN}(z_i) = \max(0,\:z_i \boldsymbol{W_{\textit{L}_1}} + b_1) \boldsymbol{W_{\textit{L}_2}} + b_2
\end{align*}
where $\boldsymbol{W_{\textit{L}_1}}\in\mathbb{R}^{d_\textit{m}\times d_\textit{h}}$, $\boldsymbol{W_{\textit{L}_2}}\in\mathbb{R}^{d_\textit{h}\times d_\textit{m}}$, $b_1$ and $b_2$ are biases, and $d_\textit{h}$ is hidden size.
The $\mathrm{FFN}$ sublayer outputs are subsequently given as input to the next encoder layer.

The decoder layer consists of three sublayers. The first sublayer, similar to the encoder, performs masked self-attention where masks are used to prevent positions from attending to subsequent positions. The second sublayer performs \emph{encoder-decoder inter-attention} where the input to the query vector comes from the decoder layer while the input to the key and value vectors comes from the encoder's last layer. To denote parameters in these two sublayers, the transformation weights of the masked self-attention sublayer are referenced as $\boldsymbol{W_{\textit{K}}^1}, \boldsymbol{W_{\textit{V}}^1}, 
\boldsymbol{W_{\textit{Q}}^1}, 
\boldsymbol{W_{\textit{F}}^1}$ and encoder-decoder attention sublayer as $\boldsymbol{W_{\textit{K}}^2}, \boldsymbol{W_{\textit{V}}^2}, 
\boldsymbol{W_{\textit{Q}}^2}, 
\boldsymbol{W_{\textit{F}}^2}$, which is also indicated in Figure~\ref{fig:tf_decoder}. The third sublayer consists of an $\mathrm{FFN}$. To generate predictions for the next word, there is a linear layer on top of the decoder layer. The weight of this linear layer is shared with the weight of the embedding layer~\cite{inan2016tying}.

Residual connections~\citep{he2016deep} and layer normalization~\citep{ba2016layer} are applied on each sublayer and to the output vector from the final encoder and decoder layers.

\subsection{Parameter Sharing Strategies} \label{subsec:proposed_method}
In this paper, our objective is to investigate effective parameter sharing strategies for the Transformer model using MTL, mainly for \emph{one-to-many} multilingual translation.
Here, we will use the symbol $\boldsymbol{\Theta}$ to denote the set of shared parameters in our model. These parameter sharing strategies are described below:
\begin{itemize}
\item The base case consists of separate bilingual translation models for each language pair \mbox{\big($\boldsymbol{\Theta}$ = $\emptyset$\big)}.

\item Use of a common embedding layer for all the bilingual models \big($\boldsymbol{\Theta}$ = \{$\boldsymbol{W_{\textit{E}}}\}$\big). This will result in a significant reduction of the total parameters by sharing parameters across common words present in the source and target sentences~\cite{wu2016google}.

\item Use of a common encoder for the source language and a separate decoder for each target language \big($\boldsymbol{\Theta}$ = \{$\boldsymbol{W_{\textit{E}}}$, $\boldsymbol{\theta}_{\textit{ENC}}\}$\big). This has the advantage that the encoder will now see more source language training data~\cite{dong2015multi}.
\end{itemize}
Next, we also include the decoder parameters among the set of shared parameters. While doing so, we will assume that the embedding and the encoder parameters are always shared between the bilingual models. Because there can be exponentially many combinations considering all the different feasible sets of shared parameters between the multiple decoders, we only select a subset of these combinations based on our preliminary results. These selected weights are shared in all the layers of the decoder unless stated otherwise. A schematic diagram illustrating the various possible parameter matrices that can be shared in each sublayer of our MTL model is shown in Figure~\ref{fig:pshare}.

\begin{itemize}
\item We share only the $\mathrm{FFN}$ sublayer parameters \mbox{\big($\boldsymbol{\Theta}$ = \big\{$\boldsymbol{W_{\textit{E}}}$, $\boldsymbol{\theta}_{\textit{ENC}}$, $\boldsymbol{W_{\textit{L}_1}}$, $\boldsymbol{W_{\textit{L}_2}}\big\}$\big)}.

\item Sharing the weights of the self-attention sublayer  
\big($\boldsymbol{\Theta}$ = \big\{$\boldsymbol{W_{\textit{E}}}$, $\boldsymbol{\theta_{\textit{ENC}}}$, $\boldsymbol{W_{\textit{K}}^1}$, $\boldsymbol{W_{\textit{Q}}^1}$, $\boldsymbol{W_{\textit{V}}^1}$, $\boldsymbol{W_{\textit{F}}^1}$\big\}\big).

\item Sharing the weights of the encoder-decoder attention sublayer \big($\boldsymbol{\Theta}$ = \big\{$\boldsymbol{W_{\textit{E}}}$, $\boldsymbol{\theta_{\textit{ENC}}}$, $\boldsymbol{W_{\textit{K}}^2}$, $\boldsymbol{W_{\textit{Q}}^2}$, $\boldsymbol{W_{\textit{V}}^2}$, $\boldsymbol{W_{\textit{F}}^2}$\big\}\big).

\item We limit the attention parameters that are shared to only include either the key and query weights \big($\boldsymbol{\Theta}$ = \big\{$\boldsymbol{W_{\textit{E}}}$, $\boldsymbol{\theta}_{\textit{ENC}}$, $\boldsymbol{W_{\textit{K}}^1}$, 
$\boldsymbol{W_{\textit{Q}}^1}$, $\boldsymbol{W_{\textit{K}}^2}$, 
$\boldsymbol{W_{\textit{Q}}^2}$\big\}\big) or the key and value weights \mbox{\big($\boldsymbol{\Theta}$ = \big\{$\boldsymbol{W_{\textit{E}}}$, $\boldsymbol{\theta}_{\textit{ENC}}$, $\boldsymbol{W_{\textit{K}}^1}$, $\boldsymbol{W_{\textit{V}}^1}$, $\boldsymbol{W_{\textit{K}}^2}$, $\boldsymbol{W_{\textit{V}}^2}$\big\}\big)}. The motivation for doing so is so that the shared attention sublayer weights can model the common aspects of the target languages while the individual $\mathrm{FFN}$ sublayer weights can model the distinctive or unique aspects of each language.

\item We share all the parameters of the decoder to have a single unified model \big($\boldsymbol{\Theta}$ = \big\{$\boldsymbol{W_{\textit{E}}}$, $\boldsymbol{\theta}_{\textit{ENC}}$, $\boldsymbol{\theta}_{\textit{DEC}}$\big\}\big). Fewer parameters in the decoder indicates limited modeling ability, and we expect this method to obtain good translation accuracy mainly when the target languages are related~\cite{johnson2017google}.
\end{itemize}
   
\section{Experimental Setup} \label{sec:exp_setup}
In this section, first, we describe the datasets used in this work and the evaluation criteria. Then, we describe the training regimen followed in all our experiments. All of our models were implemented in PyTorch framework~\cite{paszke2017automatic} and were trained on a single GPU.

\subsection{Datasets and Evaluation Metric}
\begin{table}[t]
\centering
\begin{tabular}{l|ccc}
 \toprule
 \textbf{Language Pair} & \textbf{Training} & \textbf{Dev} & \textbf{Test} \\
 \midrule
 \textsc{En}$-$\textsc{Ro} & 180,484 & 3,904 & 4,631 \\
 \textsc{En}$-$\textsc{Fr} & 192,304 & 4,320 & 4,866 \\
 \textsc{En}$-$\textsc{Nl} & 183,767 & 4,459 & 5,006 \\
 \textsc{En}$-$\textsc{De} & 167,888 & 4,148 & 4,491 \\
 \textsc{En}$-$\textsc{Ja} & 204,090 & 4,429 & 5,565 \\
 \textsc{En}$-$\textsc{Tr} & 182,470 & 4,045 & 5,029 \\
 \bottomrule
 \end{tabular}
\caption{Number of sentences in the training, dev, and test splits for each language pair used in our experiments. The languages are represented by their ISO 639-1 codes \emph{En:English}, \emph{Fr:French}, \emph{Nl:Dutch}, \emph{De:German}, \emph{Ja:Japanese}, \emph{Tr:Turkish}.}
\label{table:dataset_stat}
\end{table}

To perform multilingual translation experiments, we select six language pairs from the openly available TED talks dataset~\cite{Ye2018WordEmbeddings} whose statistics are mentioned in Table~\ref{table:dataset_stat}. This dataset already contains predefined splits for training, development, and test sets. Among these languages, Romanian (\textsc{Ro}) and French (\textsc{Fr}) are \emph{Romance} languages, German (\textsc{De}) and Dutch (\textsc{Nl}) are \emph{Germanic} languages while Turkish (\textsc{Tr}) and Japanese (\textsc{Ja}) are unrelated languages that come from distant language families. For all language pairs, tokenization was carried out using the \texttt{Moses} tokenizer,\footnote{\href{https://github.com/moses-smt/mosesdecoder/tree/master/scripts/tokenizer}{https://github.com/moses-smt/mosesdecoder/tree/master/scripts/tokenizer}} except for Japanese, where word segmentation was performed using the \texttt{KyTea} tokenizer~\cite{neubig2011pointwise}. To select training examples, we filter sentences with a maximum length of $70$ tokens. For evaluation, we report the model's performance using the standard BLEU score metric~\cite{papineni2002bleu}. We use the \texttt{mtevalv14.pl} script from the \texttt{Moses} toolkit to compute the tokenized BLEU scores.

\subsection{Training Protocols}
In this work, we follow the same training process for all the experiments. We jointly encode the source and target language words with subword units by applying \emph{byte pair encoding}~\cite{gage1994nad} with 32,000 merge operations~\cite{sennrich2016neural}. These subword units restrict the vocabulary size and prevent the need for explicitly handling out-of-vocabulary symbols as the vocabulary can be used to represent any word.
We use \emph{LeCun uniform initialization} \cite{lecun1998efficient} for all the trainable model parameters. Embedding layer weights are randomly initialized according to truncated Gaussian distribution
$\boldsymbol{W_{\textit{E}}}\sim\mathcal{N}(0,{d_{\textit{m}}}^{-1/2})$.

In all the experiments, we use \emph{Transformer base model} configuration~\cite{vaswani2017attn} that consists of six encoder-decoder layers, $d_{\textit{m}}=512$, $d_{\textit{h}}=2,048$, and $\ell=8$. For optimization, we use SGD with Adam optimizer~\citep{kingma2014adam} with $\beta_1=0.9$, $\beta_2=0.997$, and $\epsilon=1e^{-9}$.\footnote{These hyperparameter values are based on the single GPU transformer model from the open-source \emph{tensor2tensor} toolkit.} 
The learning rate ($\textit{lr}$) schedule is varied at every optimization step ($\textit{step}$) according to:
\begin{align*}
\textit{lr} = 2 \textit{d}_\textit{m}^{-0.5} \mathrm{min}\left(\textit{step}^{-0.5},\textit{step}\cdot 16000^{-1.5}\right)
\end{align*}
Each mini-batch consists of approximately $3,000$ source and $3,000$ target tokens such that similar length sentences are bucketed together. We train the models until convergence and save the best checkpoint using development set performance. 
For model regularization, we use label smoothing $(\epsilon=0.1)$~\cite{pereyra2017regularizing} and apply dropout (with $p_{drop}=0.1$)~\cite{srivastava2014dropout} to the word embeddings, attention coefficients, ReLU activation, and to the output of each sublayer before the residual connection. During decoding, we use beam search with beam width $5$ and length normalization with $\alpha=1$~\cite{wu2016google}.

\subsection{Multilingual Training}
During the multilingual model's training and inference, we include an additional token representing the desired target language at the start of each source sentence~\cite{johnson2017google}. The presence of this additional token will help the model learn the target language to translate to during decoding. For preprocessing, we apply byte pair encoding over the combined dataset of all the language pairs. We perform model training using balanced mini-batches \emph{i.e.}\ it contains roughly an equal number of sentences for every target language. While training, we compute weighted average cross-entropy loss where the weighting term is proportional to the total word count observed in each of the target language sentences.

\section{Results} \label{sec:results}
In this section, we will describe the results of our proposed parameter sharing techniques and later present the broader context by comparing them with bilingual translation models and previous benchmark methods.

\begin{table*}[t]
\begin{minipage}{\linewidth}
\centering
  \begin{tabular}{l|rr|rr|r}
  \toprule
  \multirow{2}{*}{\textit{Set of shared parameters} ($\Theta$)} &  
  \multicolumn{2}{c|}{\textsc{En}$\rightarrow$\textsc{Ro}$+$\textsc{Fr}} &       
  \multicolumn{2}{c|}{\textsc{En}$\rightarrow$\textsc{De}$+$\textsc{Nl}} & \textit{params} \\
  \cmidrule{2-5} & 
$\rightarrow$\textsc{Ro} & $\rightarrow$\textsc{Fr} & $\rightarrow$\textsc{De} & $\rightarrow$\textsc{Nl} & $\times 10^6$ \\
\midrule
$W_{\textit{E}}$ & 27.21 & 43.36 & 30.32 & 33.51 & 105 \\
$W_{\textit{E}}$, $\theta_{\textit{ENC}}$ & 27.82 & 43.83 & 29.97 & 33.33 & 86 \\
$W_{\textit{E}}$, $\theta_{\textit{ENC}}$, $W_1,\ W_2$ & 27.78 & 43.87 & 29.95 & 33.12 & 74 \\
$W_{\textit{E}}$, $\theta_{\textit{ENC}}$, $W_{\textit{K}}^1$, $W_{\textit{Q}}^1$, $W_{\textit{V}}^1$, $W_{\textit{F}}^1$ & 27.80 & 43.76 & 30.68 & 33.99 & 80 \\
$W_{\textit{E}}$, $\theta_{\textit{ENC}}$, $W_{\textit{K}}^2$, $W_{\textit{Q}}^2$, $W_{\textit{V}}^2$, $W_{\textit{F}}^2$ & 28.36 & 44.19 & 30.50  & 33.75 & 80 \\
$W_{\textit{E}}$, $\theta_{\textit{ENC}}$, $W_{\textit{K}}^1$, $W_{\textit{V}}^1$, $W_{\textit{K}}^2$, $W_{\textit{V}}^2$ & 27.77 & 43.83 & 30.54 & 34.00 & 80 \\
$W_{\textit{E}}$, $\theta_{\textit{ENC}}$, $W_{\textit{K}}^1$, $W_{\textit{Q}}^1$, $W_{\textit{K}}^2$, $W_{\textit{Q}}^2$ & 27.58 & 43.84 & \textbf{30.70} & \textbf{34.05} & 80 \\
$W_{\textit{E}}$, $\theta_{\textit{ENC}}$, $W_{\textit{K}}^1$, 
$W_{\textit{Q}}^1$, $W_{\textit{V}}^1$, $W_{\textit{F}}^1$, $W_{\textit{K}}^2$, $W_{\textit{Q}}^2$, $W_{\textit{V}}^2$, $W_{\textit{F}}^2$ 
& 28.14 & 44.12 & 30.64 & 33.92 & 74 \\
$W_{\textit{E}}$, $\theta_{\textit{ENC}}$, $\theta_{\textit{DEC}}$ & \textbf{28.52} & \textbf{44.28} & 30.45 & 33.69 & 61 \\
\bottomrule
\end{tabular}
\subcaption{The target languages in this \emph{one-to-many} translation task belong to the same language family. \textsc{Ro} and \textsc{Fr} are \emph{Romance} languages while \textsc{De} and \textsc{Nl} are \emph{Germanic} languages.}
\medskip\medskip
\label{table:O2M_simlang}
\end{minipage}
\begin{minipage}{\linewidth}
\centering
\begin{tabular}{l|rr|rr|r}
\toprule
\multirow{2}{*}{\textit{Set of shared parameters} ($\Theta$)} &  
\multicolumn{2}{c|}{\textsc{En}$\rightarrow$\textsc{De}$+$\textsc{Tr}} 
& 
\multicolumn{2}{c|}{\textsc{En}$\rightarrow$\textsc{De}$+$\textsc{Ja}} 
&
\textit{params} \\
\cmidrule{2-5}
&
$\rightarrow$\textsc{De} & $\rightarrow$\textsc{Tr} & $\rightarrow$\textsc{De} & $\rightarrow$\textsc{Ja} & $\times 10^6$
\\
\midrule
$W_{\textit{E}}$ & 30.35 & 19.66 & 30.10 & 18.62 & 105 \\
$W_{\textit{E}}$, $\theta_{\textit{ENC}}$ & 30.55 & 19.29 & 30.21 & 18.70 & 86 \\
$W_{\textit{E}}$, $\theta_{\textit{ENC}}$, 
$W_{L_1},\ W_{L_2}$ &  30.21 & 19.17 & 30.36 & 18.92 & 74 \\
$W_{\textit{E}}$, $\theta_{\textit{ENC}}$, $W_{\textit{K}}^1$, $W_{\textit{Q}}^1$, $W_{\textit{V}}^1$, $W_{\textit{F}}^1$  
  & 30.35 & 19.24 & 30.05  & 18.78 & 80 \\
$W_{\textit{E}}$, $\theta_{\textit{ENC}}$, $W_{\textit{K}}^2$, 
$W_{\textit{Q}}^2$, $W_{\textit{V}}^2$, $W_{\textit{F}}^2$ 
 & 30.49 & 19.40 & 30.16 & 18.73 & 80 \\
$W_{\textit{E}}$, $\theta_{\textit{ENC}}$, $W_{\textit{K}}^1$, $W_{\textit{V}}^1$, $W_{\textit{K}}^2$, $W_{\textit{V}}^2$ & 30.66 & 19.34 & 30.36 & 18.92 & 80 \\
$W_{\textit{E}}$, $\theta_{\textit{ENC}}$, $W_{\textit{K}}^1$, $W_{\textit{Q}}^1$, $W_{\textit{K}}^2$, $W_{\textit{Q}}^2$ & \textbf{30.71} & \textbf{19.67} & \textbf{30.48} & \textbf{19.00} & 80 \\
$W_{\textit{E}}$, $\theta_{\textit{ENC}}$, $W_{\textit{K}}^{1}$, $W_{\textit{Q}}^{1}$, $W_{\textit{V}}^{1}$, $W_{\textit{F}}^{1}$, $W_{\textit{K}}^{2}$, $W_{\textit{Q}}^{2}$, $W_{\textit{V}}^{2}$, $W_{\textit{F}}^{2}$ & 30.40 & 19.35 & 30.35 & 18.80 & 74 \\
$W_{\textit{E}}$, $\theta_{\textit{ENC}}$, $\theta_{\textit{DEC}}$ & 28.74 & 18.69 & 29.68 & 18.50 & 61 \\
\bottomrule
\end{tabular}
\subcaption{The target languages in this \emph{one-to-many} translation task belong to distant language families. \textsc{De}, \textsc{Tr}, and \textsc{Ja} are unrelated as they belong to \textit{Germanic}, \textit{Turkic}, and \textit{Japonic} language families respectively.}
\label{table:O2M_dislang}
\end{minipage}
\caption{BLEU scores for various parameter sharing strategies when the target languages either belong to the same family (\{\textsc{Ro}, \textsc{Fr}\}, \{\textsc{De}, \textsc{Nl}\}) or to distant families ({\textsc{De}, \textsc{Tr}, \textsc{Ja}}). $\theta_{\textit{ENC}}$ denotes that all the encoder parameters are shared between the models; $\theta_{\textit{DEC}}$ denotes that all the decoder parameters are shared between the models.}
\label{table:param_sharing}
\end{table*}
\subsection{Parameter Sharing}
Here, we first analyze the results of \emph{one-to-many} multilingual translation experiments when there are two target languages and both of them belong to the same language family. The first set of experiments are on \emph{Romance} languages (\textsc{En}$\rightarrow$\textsc{Ro}$+$\textsc{Fr}) and the second set of experiments are on \emph{Germanic} languages (\textsc{En}$\rightarrow$\textsc{De}$+$\textsc{Nl}). We report the BLEU scores in Table~\ref{table:O2M_simlang} when different sets of parameters are shared in these experiments. We observe that sharing only the embedding layer weight between the multiple models leads to the lowest scores. Sharing the encoder weights results in significant improvement for \textsc{En}$\rightarrow$\textsc{Ro}$+$\textsc{Fr} but leads to a small decrease in \textsc{En}$\rightarrow$\textsc{De}$+$\textsc{Nl} scores. 

We then gradually include both the decoder's weights to the set of shareable parameters. Specifically, we include the parameters of $\mathrm{FFN}$, self-attention, encoder-decoder attention, both the attention sublayers, key, query, value weights from both the attention sublayers, and finally all the parameters of the decoder layer. From the results, we note that the sharing of the encoder-decoder attention weights leads to substantial gains. Finally, sharing the entirety of the parameters (\emph{i.e.} having one model) leads to the best BLEU scores for \textsc{En}$\rightarrow$\textsc{Ro}$+$\textsc{Fr} and sharing only the key and query matrices from both the attention layers leads to the best BLEU scores for \textsc{En}$\rightarrow$\textsc{De}$+$\textsc{Nl}. One of the reasons for such large increase in BLEU is that encoder has access to more English language training data and for the decoder, as the target languages belong to the same family, they may contain common vocabulary, thus improving the generalization error for both the target languages.

Next, we analyze the results of \emph{one-to-many} translation experiments when both the target languages belong to distant language families and are unrelated. The first set of experiments are on \emph{Germanic, Turkic} languages (\textsc{En}$\rightarrow$\textsc{De}$+$\textsc{Tr}) and the second set of experiments are on \emph{Germanic, Japonic} languages (\textsc{En}$\rightarrow$\textsc{De}$+$\textsc{Ja}). We present the results in Table~\ref{table:O2M_dislang} when different sets of parameters are shared. Here, we observe that the approach of sharing all the parameters leads to a noticeable drop in the BLEU scores for both the considered language pairs. Similar to the above discussion, sharing the key and query matrices results in a large increase in the BLEU scores. We hypothesize that in this partial parameter sharing strategy, the sharing of key and query attention weights effectively models the common linguistic properties while the separate $\mathrm{FFN}$ sublayer weights model the unique characteristics of each target language, thus overall leading to a large improvement in the BLEU scores. The results of other decoder parameter sharing approaches lie close to the key and query parameter sharing method. As the target languages are from different families, their vocabularies may have some overlap but will be significantly different from each other. In this scenario, a useful alternative is to consider a separate embedding layer for every source-target language pair while sharing all the encoder and decoder parameters. However, we did not experiment with this approach, as the inclusion of separate embedding layers will lead to a large increase in the model parameters and as a result model training will become more memory intensive. We leave the investigation of such parameter sharing strategy to future work.

\subsection{Overall Comparison}

\begin{table*}[!ht]
\centering
\medskip
\begin{tabular}{l|rr|rrV{2.5}rr|rr|r}
\toprule
Method & \multicolumn{2}{c|}{\textsc{En}$\rightarrow$\textsc{De}$+$\textsc{Tr}} & \multicolumn{2}{cV{2.5}}{\textsc{En}$\rightarrow$\textsc{De}$+$\textsc{Ja}} & \multicolumn{2}{c|}{\textsc{En}$\rightarrow$\textsc{Ro}+\textsc{Fr}} & \multicolumn{2}{c|}{\textsc{En}$\rightarrow$\textsc{De}$+$\textsc{Nl}} & \textit{params}
\\
\cmidrule{2-9} & 
$\rightarrow$\textsc{De} & $\rightarrow$\textsc{Tr} & $\rightarrow$\textsc{De} & $\rightarrow$\textsc{Ja} & $\rightarrow$\textsc{Ro} & $\rightarrow$\textsc{Fr} & $\rightarrow$\textsc{De} & $\rightarrow$\textsc{Nl} & $\times 10^6$
\\
\midrule
GNMT NS & 27.01 & 16.07 & 27.01 & 16.62 & 24.38 & 40.50 & 27.01 & 30.64 & -- \\
GNMT FS & 29.07 & 18.09 & 28.24 & 17.33 & 26.41 & 42.46 & 28.52 & 31.72 & -- \\
Transformer NS & 29.31 & 18.62 & 29.31 & 17.92 & 26.81 & 42.95 & 29.31 & 32.43 & 122 \\
Transformer FS & 28.74 & 18.69 & 29.68 & 18.50 & \textbf{28.52} & \textbf{44.28} & 30.45 & 33.69 & 61 \\
Transformer PS & \textbf{30.71} & \textbf{19.67} & \textbf{30.48} & \textbf{19.00} & 27.58 & 43.84 & \textbf{30.70} & \textbf{34.05} & 80 \\
\bottomrule
\end{tabular}
\captionof{table}{BLEU scores for different models for \emph{one-to-many} translation task. \textbf{NS}: \textit{No Sharing} corresponds to the bilingual models when the two language pairs are trained independently; \textbf{FS}: \textit{Full Sharing} means one model is used for the translation of all the language pairs; \textbf{PS}: \textit{Partial Sharing} means that the embedding, encoder, decoder's key, and value weights are shared between the two models.}
\label{table:One2Many}
\end{table*}
In Table~\ref{table:One2Many}, we show an overall performance comparison of no parameter sharing, full parameter sharing for both GNMT~\cite{wu2016google} and Transformer models, and the best approaches according to maximum BLEU score from our partial parameter sharing strategies. For training the GNMT models, we use its open-source implementation\footnote{\href{https://github.com/tensorflow/nmt}{https://github.com/tensorflow/nmt}}~\cite{luong17neural} with four layers\footnote{We found that the four layer model for GNMT didn't overfit and obtained the best BLEU scores.} and default parameter settings. First, we note that the BLEU scores of the Transformer model are always better than the GNMT model by a significant margin for both bilingual (no sharing) and multilingual (full sharing) translation tasks. This reflects that the Transformer model is well-suited for both multilingual and bilingual translation tasks compared with the GNMT model. We also surprisingly note that the GNMT fully shared model is able to consistently obtain higher BLEU scores compared with its bilingual version irrespective of which families the target languages belong to.

However, for the \emph{one-to-many} translation task when the target languages are from distant families, we observe that fully shared Transformer model leads to a substantial drop or small gains in the BLEU score compared with the bilingual models. Specifically, for the \textsc{En}$\rightarrow$\textsc{De}$+$\textsc{Tr} setting, BLEU drops by 0.6 for \textsc{En}$\rightarrow$\textsc{De}, while staying even for \textsc{En}$\rightarrow$\textsc{Tr}.  In contrast, our method of sharing embedding, encoder, decoder's key, and query parameters leads to substantial increases in BLEU scores (1.4$\uparrow$ for \textsc{En}$\rightarrow$\textsc{De} and 1.1$\uparrow$ for \textsc{En}$\rightarrow$\textsc{Tr}). Similarly, for \textsc{En}$\rightarrow$\textsc{De}$+$\textsc{Ja}, using the fully shared Transformer model, we observe small gains of 0.3 and 0.5 BLEU points for \textsc{En}$\rightarrow$\textsc{De} and \textsc{En}$\rightarrow$\textsc{Ja} respectively while our partial parameter sharing method again leads to significant improvements (1.5$\uparrow$ for \textsc{En}$\rightarrow$\textsc{De} and 1.1$\uparrow$ for \textsc{En}$\rightarrow$\textsc{Ja}). This demonstrates the utility of our proposed partial parameter sharing method.

We also note that fully shared Transformer models can be an effective strategy only when both the target languages are from the same family. For the task of \textsc{En}$\rightarrow$\textsc{Ro}$+$\textsc{Fr}, the fully shared model performs surprisingly well and yields significant improvements of 1.7 and 1.3 BLEU points compared with bilingual models for \textsc{En}$\rightarrow$\textsc{Ro} and \textsc{En}$\rightarrow$\textsc{Fr} respectively. A similar increase in performance can also be observed for the \textsc{En}$\rightarrow$\textsc{De}$+$\textsc{Nl} task, although for this task, our partial parameter sharing method (encoder, embedding, decoder's key, and query weights) obtains even higher BLEU scores. (1.4$\uparrow$ for \textsc{En}$\rightarrow$\textsc{De} and 1.6$\uparrow$ \textsc{En}$\rightarrow$\textsc{Nl}).

\subsection{Analysis}
\begin{table*}[!t]
\small
\centering
\begin{tabular}{l|l}
\toprule
\textbf{source} & So half a year ago , I decided to go to Pakistan myself .\\
\textbf{reference} & Vor einem halben Jahr entschied ich mich , selbst nach Pakistan zu gehen .\\
\textbf{partial sharing} & Vor einem halben Jahr entschied ich mich , selbst nach Pakistan zu gehen . (1.0) \\
\textbf{full sharing} & Vor \underline{eineinhalb} 
\underline{Jahren} beschloss ich , nach Pakistan zu gehen . (0.35) \\
\midrule
\textbf{source} & Your heart starts beating faster .\\
\textbf{reference} & Ihr Herz beginnt schneller zu schlagen .\\
\textbf{partial sharing} & Ihr Herz beginnt schneller zu schlagen . (1.0)\\
\textbf{full sharing} & Ihr Herz \underline{schlägt} schneller . (0.27)\\
\bottomrule
\end{tabular}
\caption{Sample translations from \textsc{En}$\rightarrow$\textsc{De} when \emph{one-to-many} multilingual model was trained on unrelated target language pairs \textsc{En}$\rightarrow$\textsc{De}$+$\textsc{Tr}. In these examples, the method of partial sharing of decoder parameters obtains a very high BLEU score (mentioned in parentheses).}
\label{table:exp-translation}
\end{table*}

Here, we analyze the generated translations of the partial sharing and full sharing approaches for \textsc{En}$\rightarrow$\textsc{De} when \emph{one-to-many} multilingual model was trained on unrelated target language pairs \textsc{En}$\rightarrow$\textsc{De}$+$\textsc{Tr}. These translations were obtained using the test set of \textsc{En}$\rightarrow$\textsc{De} task. Here partial sharing refers to the specific approach of sharing the embedding, encoder, and decoder's key and query parameters in the model.

We show example translations in Table~\ref{table:exp-translation} where partial sharing method gets a high BLEU score (shown in parentheses) but the full sharing method does not. We see that sentences generated by partial sharing method are both semantically and grammatically correct while the full sharing method generates shorter sentences compared with reference translations. As highlighted in table cells, the partial sharing method is able to correctly translate a mention of relative time ``\emph{half a year}'' and a co-reference expression ``\emph{mich}''. In contrast, the fully shared model generates incorrect expressions of time mentions ``\emph{eineinhalb Jahren}'' (one and half years) and different verb forms (``\emph{schlägt}'' is generated vs ``\emph{schlagen}'' in the reference).

\begin{figure}[!ht]
\centering
\includegraphics[max width=\linewidth, scale=1.0]{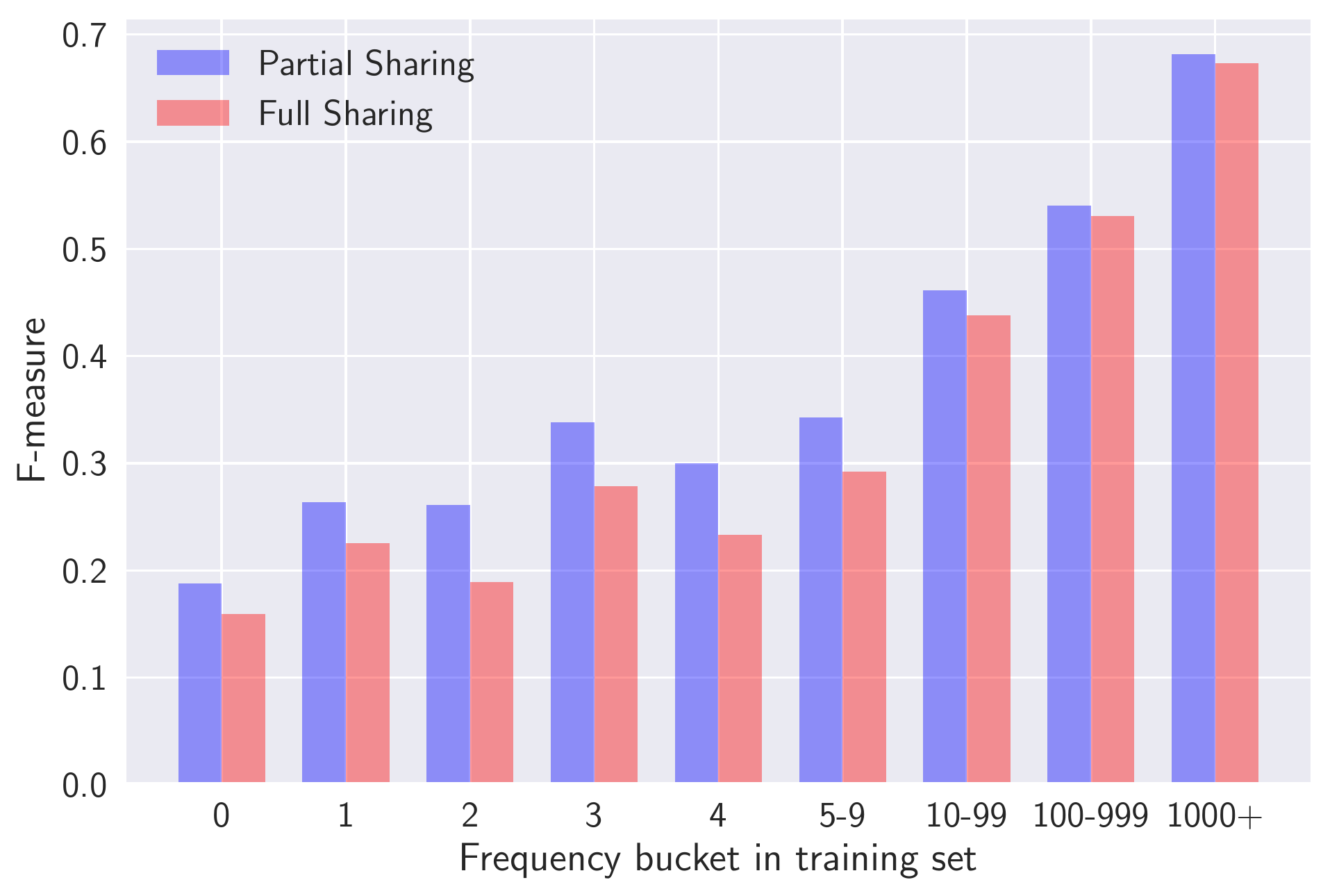}
\caption{The F-measure for the target language (\textsc{De}) words in \emph{one-to-many} multilingual translation task (\textsc{En}$\rightarrow$\textsc{De}$+$\textsc{Tr}). Best viewed in color.}
\label{fig:fscore}
\end{figure}

We also perform a comparison of the F-measure of the target words for \textsc{En}$\rightarrow$\textsc{De}, bucketed by frequency in the training set. As displayed in Figure~\ref{fig:fscore}, this shows that the partial parameter sharing approach improves the translation accuracy for the entire vocabulary, but in particular for words that have low-frequency in the dataset.

\section{Related Work} \label{sec:related_work}
In this section, we will review the prior work related to MTL and multilingual translation.

\subsection{Multi-task learning}
~\citet{ando2005mtl} obtained excellent results by adopting an MTL framework to jointly train linear models for NER, POS tagging, and language modeling tasks involving some degree of parameter sharing. Later,~\citet{collobert2011nlp} applied MTL strategies to neural networks for tasks such as POS tagging, NER, and chunking by sharing the sequence encoder and reported moderate improvements in results. Recently,~\citet{luong2016iclr_multi} investigated MTL for a tasks such as parsing, image captioning, and translation and observed large gains in the translation task. Similarly, for MT tasks,~\citet{niehues2017exploiting} also leverage MTL by using additional linguistic information to improve the translation accuracy of NMT models. They share the encoder representations to perform joint training on translation, POS, and NER tasks. 
This kind of parameter sharing approach among multiple tasks was further extended in~\citet{zaaremoodi2018neural,zaaremoodi2018adaptive} in which they also include semantic and syntactic parsing tasks and control the relative sharing of various parameters among the tasks to obtain accuracy gains in the MT task. MTL has also been widely applied to multilingual translation that will be discussed next.

\subsection{Multilingual Translation}
On the multilingual translation task,~\citet{dong2015multi} obtained significant performance gains by sharing the encoder parameters of the source language while having a separate decoder for each target language. Later,~\citet{firat2016multi} attempted the more challenging task of \emph{many-to-many} translation by training a model that consisted of one shared encoder and decoder per language and a shared attention layer that was common to all languages. This approach obtained competitive BLEU scores on ten European language pairs while substantially reducing the total parameters. Recently,~\citet{johnson2017google} proposed a unified model with full parameter sharing and obtained comparable or better performance compared with bilingual translation scores. During model training and decoding, target language was specified by an additional token at the beginning of the source sentence. Coming to low-resource language translation,~\citet{zoph-EtAl:2016:EMNLP2016} used a transfer learning approach of fine-tuning the model parameters learned on a high-resource language pair of French$\rightarrow$English and were able to significantly increase the translation performance on Turkish and Urdu languages. Recently,~\citet{jiatong2018universal} addresses the \emph{many-to-one} translation problem for extremely low-resource languages by using a transfer learning approach such that all language pairs share the lexical and sentence-level representations. By performing joint training of the model with high-resource languages, large gains in the BLEU scores were reported for low-resource languages.

In this paper, we first experiment with the Transformer model for \emph{one-to-many} multilingual translation on a variety of language pairs and demonstrate that the approach of~\citet{johnson2017google} and~\citet{dong2015multi} is not optimal for all kinds of target-side languages. Motivated by this, we introduce various parameter sharing strategies that strike a happy medium between full sharing and partial sharing
and show that it achieves the best translation accuracy.

\section{Conclusion} \label{sec:conclusion}
In this work, we explore parameter sharing strategies for the task of multilingual machine translation using self-attentional MT models. Specifically, we examine the case when the target languages come from the same or distant language families. We show that the popular approach of full parameter sharing may perform well only when the target languages belong to the same family while a partial parameter sharing approach consisting of shared embedding, encoder, decoder's key and query weights is generally applicable to all kinds of language pairs and achieves the best BLEU scores when the languages are from distant families.

For future work, we plan to extend our parameter sharing approach in two directions. First, we aim to increase the number of target languages to more than two such that they contain a mix of both similar and distant languages and analyze the performance of our proposed parameter sharing strategies on them. Second, we aim to experiment with additional parameter sharing strategies such as sharing the weights of some specific layers (\emph{e.g.}\ the first or last layer) as different layers can encode different morphological information~\cite{belinkov2017morphology} which can be helpful in better multilingual translation.

\section*{Acknowledgments}
The authors would like to thank Mrinmaya Sachan, Emmanouil Antonios Platanios, and Soumya Wadhwa for useful comments about this work. We would also like to thank the anonymous reviewers for giving us their valuable feedback that helped to improve the paper.

\bibliographystyle{acl_natbib_nourl}
\bibliography{emnlp2018}

\end{document}